\pdfoutput=1

\documentclass[11pt]{article}

\usepackage[final]{acl}
\usepackage{amsmath}
\usepackage{times}
\usepackage{latexsym}  
\usepackage{booktabs}
\usepackage{array}
\usepackage{xcolor}
\usepackage{multirow}
\usepackage{fullpage}
\usepackage[T1]{fontenc}

\usepackage[utf8]{inputenc}

\usepackage{microtype}

\usepackage{inconsolata}

\usepackage{graphicx}

\title{UMB@PerAnsSumm 2025: Enhancing Perspective-Aware \\Summarization with Prompt Optimization and \\Supervised Fine-Tuning}

\author{
    \textbf{Kristin Qi}, \textbf{Youxiang Zhu}, \textbf{Xiaohui Liang} \\
    Department of Computer Science, University of Massachusetts Boston \\
    \texttt{\{yanankristin.qi001, youxiang.zhu001, xiaohui.liang\}@umb.edu}
      }
 
\begin{document}
\maketitle
\begin{abstract}
We present our approach to the PerAnsSumm Shared Task, which involves perspective span identification and perspective-aware summarization in community question-answering (CQA) threads. For span identification, we adopt ensemble learning that integrates three transformer models through averaging to exploit individual model strengths, achieving an 82.91\% F1-score on test data. For summarization, we design a suite of Chain-of-Thought (CoT) prompting strategies that incorporate keyphrases and guide information to structure summary generation into manageable steps. To further enhance summary quality, we apply prompt optimization using the DSPy framework and supervised fine-tuning (SFT) on Llama-3 to adapt the model to domain-specific data. Experimental results on validation and test sets show that structured prompts with keyphrases and guidance improve summaries aligned with references, while the combination of prompt optimization and fine-tuning together yields significant improvement in both relevance and factuality evaluation metrics.

\end{abstract}

\begin{figure*}[h]
    \centering
    \includegraphics[width=1.01\textwidth]{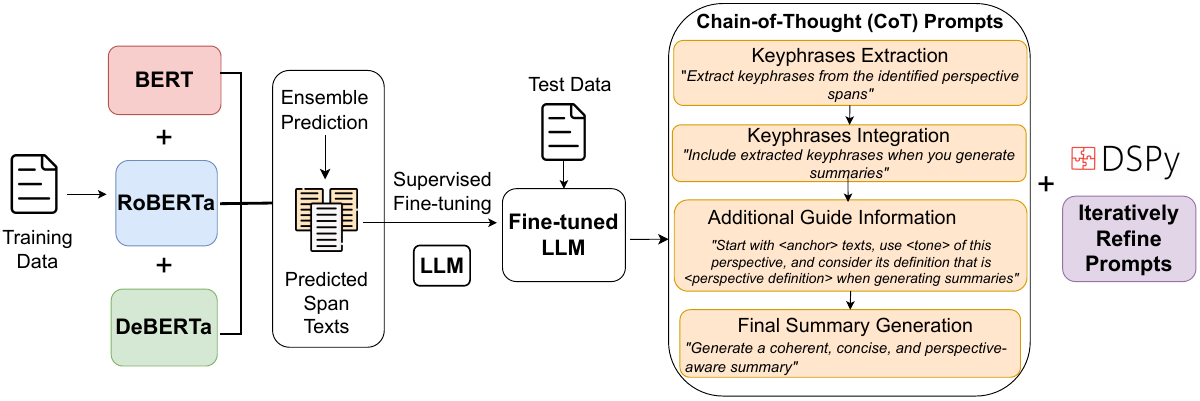}
    \caption{Detailed illustration of each component in our proposed approach for both tasks.}
    \label{fig:a_image}
\end{figure*}
\section{Introduction}
Community question-answering (CQA) platforms have transformed how medical information is exchanged, allowing users to seek and provide answers that reflect different perspectives. These responses often include general medical knowledge, personal experiences, treatment suggestions, and insights from others with similar health concerns. However, given the large volume and different viewpoints of responses presented at different locations in the answers, it is difficult to extract accurate information efficiently. Perspective-aware summarization addresses this challenge by organizing responses based on their perspectives, helping users access relevant information more effectively \cite{naik2024no}.

Recent developments in large language models (LLMs) have shown strong performance in summarization tasks. LLM-generated summaries have demonstrated comparable or superior quality to reference summaries \cite{zhang2024benchmarking, liu2023learning}. LLMs trained on medical information have enhanced their knowledge and reasoning capabilities for tackling complex problems in the healthcare domain. However, applying LLMs to perspective-aware summarization for medical CQA presents challenges: LLMs can struggle with accurately capturing distinct perspectives and effectively summarizing multiple viewpoints within long medical contexts. These challenges make it necessary to develop strategies for structuring summaries with improved accuracy.

In this work, we participate in the PerAnsSumm shared task \cite{peransumm-overview}, which focuses on developing methods for perspective span identification and perspective-aware summarization \cite{naik2024no}. Figure~\ref{fig:a_image} presents an overview of our proposed approach. For perspective span identification, we employ the ensemble learning approach that integrates three transformer-based models (BERT \cite{devlin2019bert}, RoBERTa \cite{liu2019roberta}, and DeBERTa \cite{he2020deberta}) with averaging to exploit individual model strengths and improve accuracy. For perspective-aware summarization, we leverage a pretrained LLM (Llama-3) \cite{dubey2024llama} and develop a suite of Chain-of-Thought (CoT) prompting strategies that incorporate keyphrases and additional guide information to enhance summary generation. To further improve the model performance in both relevance and factuality metrics, we apply prompt optimization using the DSPy framework \cite{khattab2023dspy} for automatic prompt refinement. We implement the 0-shot MIPRO optimizer within DSPy \cite{opsahl2024optimizing} for iterative prompt refinement. Additionally, we perform supervised fine-tuning (SFT) on Llama-3 \cite{prottasha2022transfer} to adapt the model to the domain-specific data and context-aware requirements. 

Our contributions are threefold:
\begin{itemize}
\item We integrate multiple transformer models through averaging prediction as our ensemble model. It exploits individual model strengths to achieve 82.9\% F1-score on the test set and 83.9\% on the validation set for perspective span identification.

\item We design a suite of CoT prompting approaches incorporating keyphrases and guide information to break down summarization tasks into manageable steps. To enhance summary quality, we apply DSPy automatic prompt optimization. We also implement SFT to adapt the LLM to the domain-specific data.
 
\item We conduct experiments that demonstrate the benefits of combining these approaches together. Particularly, the integration of DSPy-based prompt optimization with SFT significantly improves performance in both relevance and factuality evaluation metrics.    
\end{itemize}

\section{Related Work}
Designing and optimizing prompts have become a crucial technique for guiding LLMs to generate more accurate and relevant responses for specific tasks. Recent techniques in prompt optimization have introduced various automated strategies that are better than manual prompt engineering. These approaches leverage different techniques, including gradient-based optimization \cite{pryzant2023automatic}, reinforcement learning \cite{zhang2022tempera}, and targeted word- or phrase-level edits \cite{fernando2023promptbreeder} to automatically search for optimal prompts. The DSPy framework \cite{opsahl2024optimizing} represents an development in this direction, yielding a modular approach that enables automatic prompt refinement.

DSPy is a programming framework that allows for chaining of LLM calls through composable modules. This technique facilitates the creation of dynamic and flexible systems that can automatically optimize both prompts and weights across multiple components. DSPy enables self-refine prompts to enhance performance during inference.

The DSPy framework includes several optimizer methods specifically designed to enhance performance on downstream tasks, such as OPRO and MIPRO optimizers \cite{opsahl2024optimizing}. The OPRO optimizer leverages a stochastic mini-batch evaluation function to learn a surrogate model of the objective and refine instructions over multiple iterations. MIPRO optimizer employs a meta-optimization procedure to iteratively improve prompt construction.

Our approach applies the 0-shot MIPRO optimizer within DSPy framework to iteratively optimize instructions for generating perspective-aware summaries.

\section{Dataset and Evaluation Metrics}
\subsection{Shared Task Description}
The PerAnsSumm shared task comprises two main components that build upon each other, each addressing a different aspect of CQA.

\noindent\textbf{Perspective Span Identification:}   Detecting and labeling text spans in answers that represent each of the perspectives, including Information, Cause, Suggestion, Experience, and Question. This task requires identification of specific perspective types that appear within response texts.

\noindent\textbf{Perspective-aware Summarization:}  Generating summaries that preserve and reflect the identified perspectives and their span texts. This task creates summaries that are perspective-aware.

\subsection{Dataset}
The task dataset consists of CQA threads from medical forums \cite{naik-etal-2024-perspective}. For each thread, responses contain multiple perspectives and summaries annotated for medical question-answer pairs. The dataset is divided into three parts: train, validation, and test. The training and validation sets are provided for model development, while the test set remains hidden. The training set contains labeled CQA threads with annotated perspective spans and reference summaries, while the validation set provides additional labeled data for hyperparameter tuning.

\begin{figure}[h]
\includegraphics[width=1.03\columnwidth]{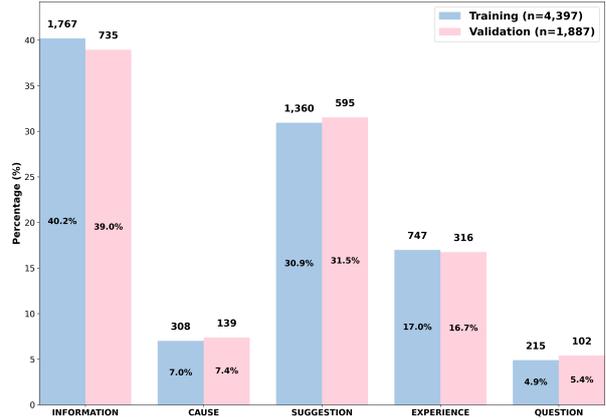} 
    \caption{Percentage distribution of each perspective type in the training and validation sets. The values displayed on top of each bar represent the actual counts.}
    \label{fig:graph}
\end{figure}

The training set contains 2236 samples, and the validation set contains 959 samples. Figure \ref{fig:graph} illustrates the distribution of each perspective type percentage in training and validation sets. Train and validation sets have a consistent percentage distribution of each perspective. 
\subsection{Evaluation Metrics} 
\textbf{Perspective-specific metrics} include the macro-averaged F1-score to evaluate classification accuracy. Strict-matching and proportional matching scores assess the similarity between predicted and reference spans.

\noindent\textbf{Summarization metrics} include two aspects: relevance and factuality. Relevance evaluation metrics include ROUGE scores (ROUGE-1, ROUGE-2, ROUGE-L) to measure n-gram overlap, BERTScore to measure semantic similarity through embeddings, and BLEU and Meteor to evaluate precision and recall of generated summaries against references. The factuality evaluation metrics use AlignScore and SummaC. AlignScore checks whether all information in the summary is in the reference. SummaC measures factual consistency between the generated and reference summaries.

\section{Method}
This section describes details of our approach to addressing the shared task: ensemble learning for span identification and prompting strategies for summarization generation, including CoT, DSPy framework, and SFT.

\subsection{Span Prediction with Ensemble Learning}
We implement an ensemble learning framework that integrates multiple transformer models. Rather than relying on a single model's prediction, ensemble learning combines predictions from multiple models to achieve better results than any single model that could attain independently.  

Our ensemble model implements three pre-trained transformer models, and we use their base models: BERT\footnote{\url{https://huggingface.co/bert-base-uncased}}, RoBERTa\footnote{\url{https://huggingface.co/FacebookAI/roberta-base}}, and DeBERTa\footnote{\url{https://huggingface.co/microsoft/deberta-base}}. These models have demonstrated strong performance in various language-related tasks. During inference, the ensemble model computes predictions through averaging that accounts for individual model predictions. Ensemble model is formally defined as:
\begin{equation}
P_{\text{ensemble}}(y \mid x) = \frac{1}{k} \sum_{i=1}^{k} P_i(y \mid x)
\end{equation}

\noindent where \( P_i(y \mid x) \) represents the prediction probability of the \( i \)-th model, and the final ensemble prediction is obtained by averaging the predictions of all \( k \) models.

\section{CoT for Summarization}
We leverage CoT prompting to enhance the reasoning and problem-solving capabilities of LLMs through breaking down the summarization task into smaller sequences of manageable steps. This approach guides the model to maintain high perspective alignment and summarization accuracy.

Our CoT prompting suite incorporates a structured four-step process:
\begin{enumerate}
  \item \textbf{Keyphrase extraction:}  We first prompt the LLM to identify and extract keyphrases from the identified perspective spans. This step elicits intermediate reasoning steps in the CoT.

 \item \textbf{Keyphrase integration:}  We prompt LLM to incorporate these extracted keyphrases when generating summaries. This step ensures that LLM preserves key information from the perspective span context. 
 
 \item \textbf{Guide information integration:}   Our prompt incorporates a set of guide information referred to as the ``guide'' in our experiments. Following the prompt design templates established in \texttt{PLASMA} \cite{naik2024no}, our guide consists of three parts:
\begin{itemize}
 \item \textbf{Tone:} Perspective-specific tone instructions (e.g., informative tone for ``Information'', understanding-seeking tone for ``Question''). \\
 \item \textbf{Anchor text:} Common start phrases found in reference summaries (e.g., \textit{``For information purposes...''} for ``Information'' and \textit{``It is inquired...'' for ``Question'')}. \\
 \item \textbf{Perspective definition:} Concise descriptions of each perspective's purpose and characteristic features.
\end{itemize}

The model is prompted to integrate guide information using the format: \textit{``Start with <anchor> texts. Use the <tone> tone of this perspective. Consider the following definition when generating the summary: <perspective definition>.''}
 
 \item \textbf{Summary generation:} Finally, we prompt the LLM to generate a coherent, concise, and perspective-aware summary: \textit{``Focus on <perspective>-specific aspects in your summary. Now generate a concise and coherent summary.''}
\end{enumerate}
The prompt template details are shown in Appendix~\ref{sec:appendix-template}. The above generation process is formalized as:
\begin{equation}
P_{\text{CoT}}(S \mid x, K, p) = \prod_{t=1}^{T} P(s_t \mid x, s_{<t}, K, p)
\end{equation}
where $x$ is the input text, $K$ represents extracted keyphrases. $p$ is the guide set for each perspective type. $S = \{s_1, s_2, \ldots, s_T\}$ represents the sequence of reasoning steps.

\subsection{Prompt Optimization with DSPy}
To further enhance summarization quality, we implement prompt optimization using the DSPy framework, which enables iterative refinement of prompts based on the context of each step. In each iteration, the DSPy compiler automatically generates multiple prompt variants (3-5) and selects optimal candidates through Bayesian optimization over the joint metric space. The challenge is defining a downstream metric that can enhance performance without having access to module-level labels or gradients.

Our downstream metric aims to balance each of the relevance evaluation metrics. Specifically, we define a composite metric that assigns equal weight (0.25) to each of four sub-metrics in the relevance category: ROUGE-L, BLEU, Meteor, and BERTScore. This process dynamically synthesizes prompts conditioned on the current step's context. The selection of the weights is based on the assumption that each sub-metric contribution is equal. The optimization objective is written as follows:

\begin{equation}
\begin{aligned}
\mathcal{L}(T) = &\, 0.25 \times \text{ROUGE-L} + 0.25 \times \text{BLEU} \\
& + 0.25 \times \text{Meteor} + 0.25 \times \text{BERTScore}
\end{aligned}
\end{equation}

The objective function of optimization can be formulated as:

\begin{equation}
\mathcal{L}(T) = 0.25 \cdot \sum_{j=1}^{n} \log P(c_j \mid x, c_{<j}, \mathcal{M}(T))
\end{equation}

\noindent where $\mathcal{L}(T)$ is the optimization objective to be maximized. $c_j$ is the generated summaries at step $j$, $\mathcal{M}(T)$ represents the LLM conditioned on optimized prompt $T$, and $P(c_j \mid x, c_{<j}, \mathcal{M}(T))$ is the probability of generating the next component $c_j$ based on prior knowledge.

\noindent \textbf{Optimizer: }  We select 0-Shot MIPRO, which provides a straightforward approach for optimizing instructions based on our balanced metric while remaining cost-effective within our computational budget constraints.

\subsection{Supervised Fine-Tuning} 
SFT on LLMs has demonstrated its success in improving performance in various domains. We implement SFT on the Llama-3-8B-Instruct model\footnote{\url{https://huggingface.co/meta-llama/Meta-Llama-3-8B-Instruct}} (Llama-3) using the Low-Rank Adaption (LoRA) technique \cite{hu2022lora}. We fine-tune the model for two epochs on the training set.  We report the results of summarization on both validation and test sets. 

\noindent\textbf{Implementation Details:}   All experiments were conducted on an NVIDIA A100 GPU with 40GB memory. We used a learning rate of $1e^{-4}$ with the AdamW optimizer and a batch size of 32. Token size was set to 256, temperature was at 0.1, and seed was at 42.

\section{Results}
We conduct all experiments using the Llama-3 model. Table \ref{tab:model-performance} presents the results for span identification, while Table \ref{tab:summarization-results} presents the results for summarization.

\section{Performance of Ensemble Models on Span Identification}
We evaluate the performance of three individual transformer models (BERT, RoBERTa, DeBERTa) and their ensemble integration. Ensemble model exploits the strengths of individual models on different evaluation metrics. Table \ref{tab:model-performance} presents comparisons on the validation set using three evaluation metrics: macro F1-score, strict match F1-score, and proportional match F1-score. The results on the test set are our final submission.

The ensemble model achieves an F1-score of 82.9\% on the test set and 83.9\% on the validation set. These results are between the best-performing (RoBERTa) and worst-performing (BERT) models. Additionally, we observe that different models outperform in different aspects of metrics: RoBERTa achieves the highest strict match F1-score, while DeBERTa performs better in proportional match.  These results indicate how individual models can outperform in an evaluation while underperforming in others, which supports the ensemble methods that could combine strengths from multiple models. Our results could be further improved through advanced ensemble techniques, such as weighted combination strategies or hierarchical model structures. 
\begin{table}[h]
\centering
\small
\begin{tabular}{l c c c}
\toprule
\textbf{Model} & \textbf{F1} & \textbf{Strict Match F1} & \textbf{Prop. Match F1} \\
\midrule
\multicolumn{4}{c}{\textit{Validation Set}} \\
\midrule
BERT & 0.813& 0.096& 0.514\\
RoBERTa & \textbf{0.858}& \textbf{0.154}& 0.546\\
DeBERTa & 0.845& 0.110& \textbf{0.559}\\
\textbf{Ensemble} & \underline{0.839}& \underline{0.120}& \underline{0.540}\\
\midrule
\multicolumn{4}{c}{\textit{Test Set Submission}} \\
\midrule
\textbf{Ensemble} & 0.829 & 0.120 & 0.505 \\
\bottomrule
\end{tabular}
\caption{Comparison of span identification performance on the validation and test sets. \textbf{Bold values} indicate the best scores, while \underline{underscored} values show results from the ensemble model.}
\label{tab:model-performance}
\end{table}
\begin{table*}[!h]
\centering
\small
\begin{tabular}{l c c c c c c c c}
\toprule
\textbf{Category}  & \textbf{R-1} & \textbf{R-2} & \textbf{R-L} & \textbf{BLEU} & \textbf{Meteor} & \textbf{BERTScore} & \textbf{AlignScore} & \textbf{SummaC} \\
\midrule
\multicolumn{9}{c}{\textit{Baseline}} \\
\midrule
Vanilla Prompting & 0.229 & 0.078 & 0.290 & 0.068 & 0.250 & 0.782 & 0.280 & 0.225 \\
\midrule
\multicolumn{9}{c}{\textit{Chain-of-Thought (CoT) Prompting (Validation Set)}} \\
\midrule
CoT\_keyphrase & 0.310& 0.110& 0.315 & 0.074 & 0.268 & 0.797 & 0.300& 0.238 \\
CoT\_guide & 0.318 & 0.108 & 0.328 & 0.081 & 0.290& 0.805 & 0.315& 0.247 \\
\midrule
\multicolumn{9}{c}{\textit{Prompt Optimization (DSPy)}} \\
\midrule
CoT\_guide+DSPy & \textbf{0.390} & \textbf{0.212}& 0.346& 0.091& 0.328& 0.830 & \textbf{0.370}& \textbf{0.291}\\
\midrule
\multicolumn{9}{c}{\textit{Supervised Fine-Tuning (SFT)}} \\
\midrule
SFT+CoT\_guide+DSPy & \textbf{0.390} & 0.165& \textbf{0.420} & \textbf{0.096} & \textbf{0.351}& \textbf{0.839}& 0.366 & 0.251\\
\midrule
\multicolumn{9}{c}{\textit{Test Set Submission Results}} \\
\midrule
SFT+CoT\_guide+DSPy & 0.360 & 0.155 & 0.328 & \textbf{0.096} & 0.339 & 0.823 & 0.333 & 0.256 \\
\bottomrule
\end{tabular}
\caption{Performance comparison of different strategies for summarization on the validation and test sets. \textit{NOTE: CoT\_guide indicates CoT+keyphrases+guide information.}}
\label{tab:summarization-results}
\end{table*}
 
\section{Summarization Performance}
We experiment with multiple prompting strategies, including vanilla prompting, CoT, DSPy-based prompt optimization, and SFT. Table \ref{tab:summarization-results} presents the comparison across eight evaluation metrics. The test set performance is our final submission. 

Our baseline uses vanilla prompting, where we directly prompt the LLM to generate concise and coherent summaries. Building on this, CoT approach with integration of keyphrases and guide information increases ROUGE-1 by +8.1\% and BERTScore by +1.5\%. These results indicate that structured reasoning and providing task-relevant external context can better guide LLM toward generating summaries with improved accuracy.

\noindent\textbf{DSPy Optimization Impact:} The application of DSPy optimization to the CoT+keyphrases+guide (\textit{CoT\_guide}) prompt strategy significantly improves performance. The DSPy framework iteratively refines prompts, leading to an additional increase across all relevance metrics (R-1, R-2, R-L, BLEU, Meteor, BERTScore), with average improvements of +25.6\% on validation set and +9.1\% on test set. Factuality metrics also show substantial improvements, with AlignScore and SummaC increasing by +17.0\% and +4.7\%, respectively. These results demonstrate that automated prompt optimization builds effectively on manual CoT design, and it scales summary quality through refinement of prompt precision and contextual awareness.

\noindent\textbf{SFT impacts:}  Fine-tuning Llama-3 using domain-specific data further enhances the model's performance when combined with DSPy optimization. On the validation set, the SFT+DSPy combination improves performance over DSPy alone, with ROUGE-L improving by +21.4\%, Meteor by +7.0\%, and BLEU by +8.8\%. Test set results reveal increases of +3.4\% for Meteor and +5.5\% for BLEU. While SFT substantially improves relevance metrics, its impact on factuality metrics is less effective, suggesting that fine-tuning primarily enhances the model's ability to generate content that aligns with reference summaries rather than improving factuality scores.

\noindent\textbf{Findings:}    We observe that combining DSPy optimization with SFT demonstrates the benefits of integrating both approaches. Fine-tuning helps the Llama-3 model adapt to domain-specific features in medical CQAs, while DSPy optimization refines the prompt structure to better guide the model's summarization. This combination particularly achieves a better performance in relevance.

\section{Limitations}
Our approach reveals several limitations. First, we use Llama-3 as our LLM without benchmarking against API-based models such as GPT-4 or Claude-3. Compared with other teams' submissions, it indicates that Llama-3 underperforms relative to GPT-4 and Claude-3. Second, our implementation of MIPRO optimizer within the DSPy framework relies on the balanced metric formulation derived from empirical assumptions. This equal-weight approach may oversimplify the relationships between different evaluation metrics and potentially reduce accuracy. The generalizability of our prompt optimization strategy also remains an open question. Alternative optimizers, such as MIPRO with bootstrapped demonstrations or OPRO may yield further improvements. Lastly, our prompt design is tailored to the medical CQA. The prompt templates do not account for potential variability within summaries. These suggest room for future research. 

\section{Conclusions}
In this paper, we present our approach to the PerAnsSumm Shared Task. Our approach adopts ensemble learning with averaging individual model predictions for span identification, achieving an 82.9\% F1-score on test data. For summary generation, we develop structured Chain-of-Thought (CoT) prompting with keyphrases and guide information and combine it with DSPy-based prompt optimization and supervised fine-tuning (SFT) of the Llama-3 model to improve summary quality. 

Our experimental results demonstrate that the integration of keyphrases and guide information within CoT improves the alignment between generated summaries and references. Notably, automated prompt optimization through the DSPy framework substantially improves both relevance and factuality evaluation metrics, with average improvements of +25.6\% on validation set. This reveals the effectiveness of iterative prompt refinement. Furthermore, combining DSPy optimization with SFT further enhances model performance, with particularly improvements in relevance metrics (ROUGE-L: +21.4\%, Meteor: +7.0\%, BLEU: +8.8\%). Future work will compare our approach with other LLMs such as GPT-4 to identify factors that impact summarization quality. Moreover, we will explore designs for metric-based optimization strategies to improve alignments with references.

\bibliography{custom}

 \appendix
\section{Prompt Template}
\label{sec:appendix-template}
\begin{figure}[h]
    \centering
    \includegraphics[width=\linewidth]{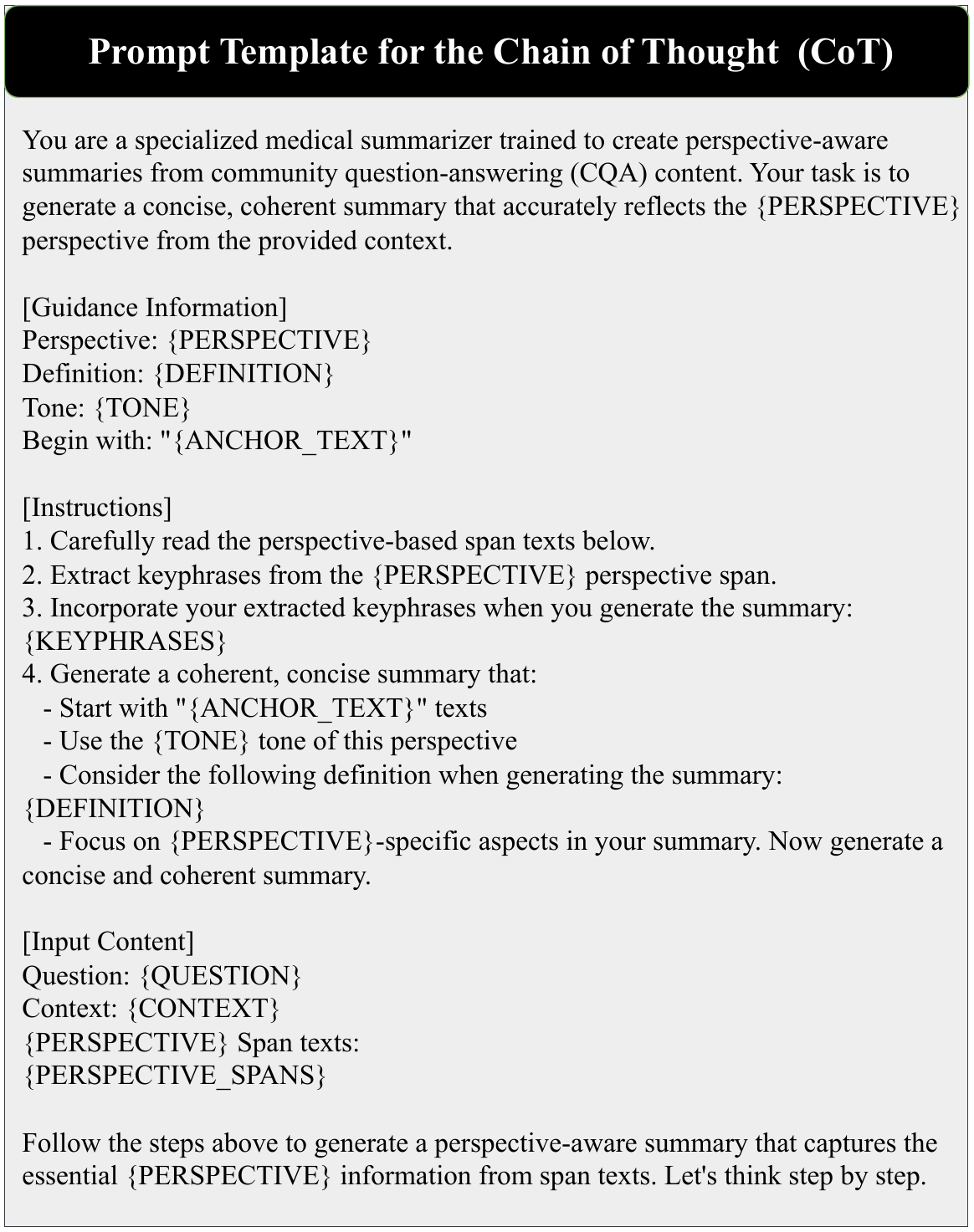}
    \caption{Prompt template used in our approach.}
    \label{fig:full-template}
\end{figure}

\end{document}